\pdfoutput=1

\documentclass[11pt]{article}

\usepackage[preprint]{acl}

\usepackage{times}
\usepackage{latexsym}

\usepackage[T1]{fontenc}

\usepackage[utf8]{inputenc}

\usepackage{microtype}

\usepackage{inconsolata}

\usepackage{booktabs}
\usepackage{makecell}
\usepackage{amsmath}
\usepackage{graphicx}
\usepackage{amsfonts}
\usepackage{fancyhdr}
\usepackage{balance}

\title{BaichuanSEED: \underline{S}haring the Potential of \underline{E}xtensiv\underline{E} \underline{D}ata Collection and Deduplication by Introducing a Competitive Large Language Model Baseline}

\author{
  \textbf{Guosheng Dong}\textsuperscript{1*} 
  \textbf{Da Pan}\textsuperscript{1} 
  \textbf{Yiding Sun}\textsuperscript{1,2$\dagger$} 
  \textbf{Shusen Zhang}\textsuperscript{1} 
  \textbf{Zheng Liang}\textsuperscript{1}  
  \\
  \textbf{Xin Wu}\textsuperscript{1} 
  \textbf{Yanjun Shen}\textsuperscript{1} 
  \textbf{Fan Yang}\textsuperscript{1} 
  \textbf{Haoze Sun}\textsuperscript{1} 
  \textbf{Tianpeng Li}\textsuperscript{1} 
  \textbf{Mingan Lin}\textsuperscript{1}  
  \\ 
  \textbf{Jianhua Xu}\textsuperscript{1}  
  \textbf{Yufan Zhang}\textsuperscript{1}  
  \textbf{Xiaonan Nie}\textsuperscript{1}  
  \textbf{Lei Su}\textsuperscript{1}  
  \textbf{Bingning Wang}\textsuperscript{1}  
  \\
  \textbf{Wentao Zhang}\textsuperscript{3} 
  \textbf{Jiaxin Mao}\textsuperscript{2} 
  \textbf{Zenan Zhou}\textsuperscript{1*} 
  \textbf{Weipeng Chen}\textsuperscript{1} 
  \\
  \textsuperscript{1}Baichuan Inc. 
  \textsuperscript{2}Gaoling School of Artificial Intelligence, Renmin University of China \\ 
  \textsuperscript{3}Peking University
  \\
 \small{
   \textbf{Correspondence:} \{\href{mailto:dongguosheng@baichuan-inc.com}{dongguosheng}, \href{mailto:zhouzenan@baichuan-inc.com}{zhouzenan}\}@baichuan-inc.com
 }
}

\begin{document}
\maketitle

\let\thefootnote\relax\footnotetext{$\dagger$ This work is finished during internship in Baichuan Inc.}
\let\thefootnote\relax\footnotetext{* Corresponding Authors.}

\begin{abstract}


The general capabilities of Large Language Models (LLM) highly rely on the composition and selection on extensive pretraining datasets, treated as commercial secrets by several institutions. 
To mitigate this issue, we open-source the details of a universally applicable data processing pipeline and validate its effectiveness and potential by introducing a competitive LLM baseline\footnote{https://baichuanseed.github.io/}. 
Specifically, the data processing pipeline consists of broad collection to scale up and reweighting to improve quality. 
We then pretrain a 7B model BaichuanSEED with 3T tokens processed by our pipeline without any deliberate downstream task-related optimization, followed by an easy but effective supervised fine-tuning stage. 
BaichuanSEED demonstrates consistency and predictability throughout training and achieves comparable performance on comprehensive benchmarks with several commercial advanced large language models, such as Qwen1.5 and Llama3. 
We also conduct several heuristic experiments to discuss the potential for further optimization on downstream tasks, such as mathematics and coding.

\end{abstract}

\section{Introduction}

Large Language Models~(LLMs) represented by ChatGPT and GPT-4~\cite{gpt4} have demonstrated exceptional performance across various domains, benefiting from the general capabilities through extensive pre-training on diverse datasets. 
Pre-training a foundation model is costly, typically requiring training on terabytes of tokens over millions of GPU hours, making the process expensive and difficult to reproduce~\cite{gpt3,palm,yulan,baichuan,qwen2,deepseek-llm,llama3-herd}. 
Many existing works focus on data selection, selecting high-quality data from massive datasets, to decrease the loss, thereby reducing computational costs and even enhancing downstream task performance~\cite{zhao2023survey,data-selection-survey}. 
These methods mainly include heuristic rule-based selection~\cite{Yulan-GARDEN,data-juicer}, model-based selection~\cite{doremi,dsdm}, dependent or relevant document detection and concatenation~\cite{skill-it,in-context-pretraining}.

Although data selection shows considerable potential, the high computational cost makes it challenging to scale up.
Additionally, the preliminary steps of data selection, data collection and reweighting are often treated as commercial secrets by companies~\cite{llama2,qwen2}, and even many open-source models that have released checkpoints do not disclose the details, hindering research advancement in this area.
Additionally, many commercial models even over-optimize to excel in specific benchmarks~\cite{llm-cheater-yutao,benchmark-leakage,benchmark-contamination-naacl2024}, thus masking their true capabilities. 
Therefore, we believe that training a transparent LLM without any specific optimization, and publicly sharing all its details will be beneficial for the community to understand the potential of data.
To emphasize, BaichuanSEED is not a SOTA model. 
What really matters to us is that understanding of a pure model helps in recognizing its actual strengths and weaknesses, the preliminary for evaluating the real impact of various optimization strategies.

In this technical report, we first introduce a 7B foundation model, BaichuanSEED, with the similar model architecture of Baichuan2~\cite{baichuan} on 3T bilingual tokens from scratch and then supervised fine-tune it to obtain BaichuanSEED-SFT, a chat model. 
Following Falcon~\cite{falcon-refinedweb}, the pre-training data processing procedure was defined to involve extensive collection to scale up, followed by reweighting data points to set the sampling probability during pre-training. 
Specifically, we collect high-quality data widely available on the internet while intentionally excluding synthetic and downstream benchmarks to ensure model purity.
Subsequently, we design a global multi-granularity deduplication algorithm to adjust the sampling weight of each data point during training. 
Here, we avoid fine-grained data selection intentionally, not to undermine its benefits but to focus more on exploring the achievable limit through data collection and reweighting. 
We then conduct a straightforward yet effective fine-tuning process to endow the model with instruction-following capabilities.

In evaluation, BaichuanSEED exhibits consistency and predictability, indicating the robustness of its training process. 
Consistency is reflected in the uniform trends observed across pre-training and fine-tuning benchmarks as the pre-training. 
Predictability refers to the ability to forecast the model's future performance based on early checkpoints. 
We also evaluate our model against a series of LLMs of similar scale on comprehensive benchmarks and several downstream tasks. 
The experimental results indicate that without excessive optimization, our model can already achieve performance comparable to advanced commercial models like Llama3 and Qwen-1.5 on several comprehensive benchmarks, while our model still has room for improvement on some downstream tasks, especially on math.
Finally, we explore the extensive potential of our model by utilizing several common optimization methods, such as adjusting the ratio of high-knowledge-density data and optimizing for mathematical and coding abilities.
We leave integrating these optimization methods into our model BaichuanSEED to construct a highly robust LLM for future work.

Our main contributions are two-fold:
(1) We propose a data processing pipeline, including broad collection to scale up and reweighting to deduplicate and improve the data quality. 
(2) We train a competitive 7B LLM baseline from scratch with 3T data processed by the aforementioned pipeline, followed by a simple yet effective supervised fine-tuning. Our model is consistent and predictable, and achieves comparable performance on comprehensive benchmarks with cutting-edge commercial LLMs without any deliberate optimization.

\section{Model Architecture}

BaichuanSEED first pre-train from scratch, followed by a Supervised Fine-Tuning~(SFT) stage for alignment.
Our model follows a Transformer decoder stack architecture similar to our previous version model Baichuan2~\cite{baichuan} and Llama~\cite{llama,llama2}. 
Specifically, our model comprises $32$ layers with $32$ attention heads.
The hidden dimension size is $4\mathrm{,}096$, and the feed-forward layer size is $11\mathrm{,}008$. 
SwiGLU~\cite{swiglu} is used as the activation function, while RMSNorm~\cite{rmsnorm} is employed to enhance training stability.
Rotary Positional Embedding~(RoPE)~\cite{rope} is employed to model relative position dependencies.

\section{Pre-training}

In this section, we first provide a detailed overview of our efforts on pre-training data. 
We follow the pipeline of first broad collection from trusted sources to scale up, then re-weighting the data to obtain diverse and high-quality pre-training data. 
We then introduce the details of the training setups.

\subsection{Pre-training Data}

The principle for constructing our pre-training data encompasses two aspects: \textbf{diversity} and \textbf{high-quality}. 
In terms of diversity, we argue that pre-training data should cover a wide range of topics, linguistic styles and formats to ensure the model can adapt to diverse application scenarios, helping it learn comprehensive world knowledge and linguistic patterns. 
Regarding high-quality, we base on the guideline that documents with high production costs typically have higher quality, which may generally curated after a rigorous process of human inspection and correction.
Additionally, we address that a healthy data type distribution across the entire dataset should be ensured, while reducing information redundancy and the proportion of low-quality data, to increase the knowledge density.

To achieve the goal of diversity and high quality, our approach focuses on both scaling up and re-weighting the documents. 
Specifically, we first collect all accessible data from trusted sources to scale up via our self-constructed pipeline. 
Then, we employ a global multi-granularity data deduplication strategy to re-weight the documents, enhancing the data quality while filtering out personal identifiable information~(PII) and low-quality data.
Following the aforementioned principles, we obtain over 10T tokens of pre-training data consequently.

\subsubsection{Collection}
\label{data-collection}

Our collected data mostly originate from the public internet, including web pages, knowledge intensive data, code, and vertical domain data.
To continuously collect the latest data for model iterations, a scraping and extracting pipeline is constructed to fetch up-to-date data automatically, enhancing the generalizability of our model in practice. 
In this section, we will introduce the details of data collection and extraction.

\noindent\textbf{Web Pages.}\quad
We collect 94 publicly available batches of CommonCrawl\footnote{https://commoncrawl.org/} spanning the past decade. 
We construct our WARC extraction and processing pipeline, considering the effectiveness and cost of the WET and WARC formats.
Trafilatura~\cite{Trafilatura}, a Python package and command-line tool designed to gather text from the web, is opted for further processing, in terms of the performance and cost among open-source web page parsing tools. 
CLD3\footnote{https://github.com/google/cld3} is utilized for document language identification.



\begin{figure}
    \centering
    \includegraphics[width=\linewidth]{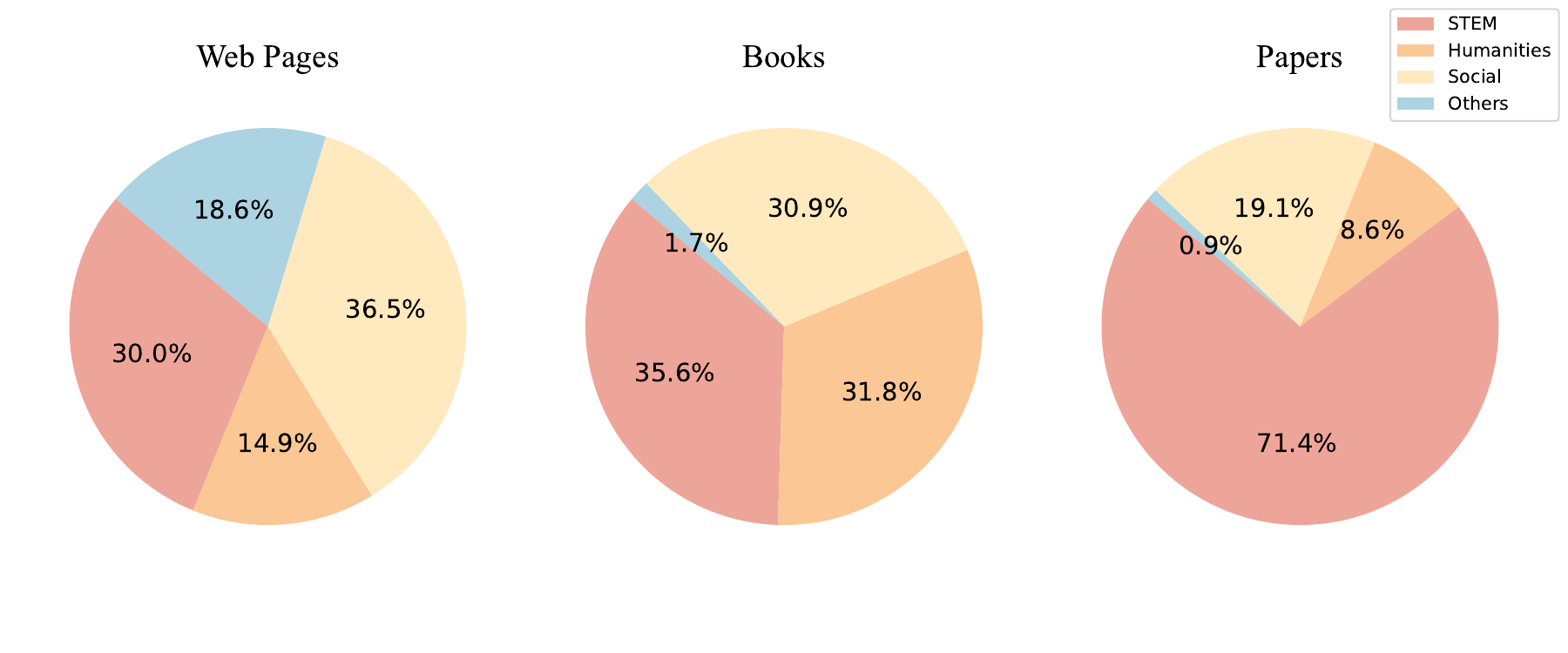}
    \caption{The detail data proportion of subjects ranging from STEM, mathematics, social science, and others, with respect to the web pages, books, and papers.}
    \label{fig:classification}
\end{figure}

\noindent\textbf{Knowledge Intensive Data.}\quad 
Apart from web pages, we further focus on Knowledge Intensive Data~(KID), including books, academic papers, and technical reports. 
KID originates from diverse domains, and possesses a more balanced distribution compared to web pages. 
As illustrated in Figure~\ref{fig:classification}, most web pages are in the domain of social science, with STEM comprising only 30\%, whereas STEM in books and papers accounts for 35\% and 72\%, respectively. 
However, processing KID is costly, due to its typical PDF format and the high requirements for document parsing. 
Consequently, we employ a hybrid approach, utilizing both the open-source tool Nougat~\cite{Nougat} and business services, to maintain the original reading order and ensure the recognition and extraction accuracy of text, tables, and formulas. 
Tables and formulas are represented in LaTeX.
We further discuss the effect of KID in Section~\ref{sec:effect-of-hkd}.

\noindent\textbf{Code.}\quad 
Our code data consist of StarCoder~\cite{starcoder}, GitHub filtered by license permissions, and StackExchange, in an interleaving form of code fragments and discussion texts. 
Both document-level and repository-level organization methods are adopted to ensure data diversity. 
For document-level, we apply the Minhash deduplication~\cite{minhash} and then use heuristic rules to select high-quality code. 
For repository-level, following Deepseek-coder~\cite{deepseek-coder-v1}, we calculate the dependencies among files within the same repository, perform topological sorting and concatenate the files. 
One repository is treated as a single sample and perform Minhash deduplication to preserve repository integrity. 
Repository-level data also undergoes heuristic rule-based pipeline to remove low-quality code and the majority of HTML, XML, JSON, and YAML files.
We also prioritize sampling Github repositories with top stars. 
Repository-level data constitutes about 62\% of the total 750B code mentioned in Table~\ref{tab:data-mixture}. 
Pivot experiments demonstrate that the mixture of repository-level and document-level data yields better model performance compared to using either one alone.

\subsubsection{Reweighting}
\label{sec:reweighting}

Reweighting is a crucial stage in our pre-training data processing strategy, affecting the sampling proportion of each data point.
We first introduce the deduplication operator, deciding whether to downsample a data point to zero.
Then a reasonable data mixture is confirmed, to balance the distribution of data from each domain.
The details of deduplication and data mixture are introduced in the following parts.

\noindent\textbf{Deduplication.}\quad 
Deduplication is the first step of reweighting.
Taking web pages as an example, we empirically found data with higher frequency is more likely tend to lead to a negative impact on foundation models.
Therefore, we propose a global multi-granularity deduplication strategy, which will reduce the amount of raw data drastically.
Via this strategy, we filter 88\% of the tokens of our CommonCrawl~(CC) dataset.
We present the details of our global multi-granularity deduplication strategy as follows, including document-level deduplication, sentence-level deduplication across documents, PII and harmful content filtering.

\begin{itemize}
    \item \textbf{Document-level Deduplication Globally.} We take both the frequency and quality into account. Specifically, existing works split CC into several batches, deduplicate in each batch, and finally merge. While we perform document-level deduplication globally, minimizing the effect of data points with extremely high frequency. We perform exact match deduplication by utilizing the MD5 value as the key of each document. Then MinHash method is adopted for similar neighbourhood document deduplication. We keep the document with the longest text length for a cluster of similar documents.
    \item \textbf{Sentence-level Deduplication across Documents.} Similar to document-level, we empirically remove sentences with extremely high frequency across documents, which mostly are meaningless junk data. For example, it may be website template breadcrumbs, standard article openings and closings, and quoted passages. These frequently recurring fixed patterns may harm the model empirically. 
    To eliminate biases in the raw data while ensuring diversity, we first split the documents into sentences, and then utilize Minhash with searched optimal hyperparameters at the sentence level across documents, making it more suitable for deduplication of short neighboring sentences. 
    Several cases after deduplication in sentence-level can be found in Appendix~\ref{sec:case-study-deduplication-appendix}.
    \item \textbf{PII and Harmful Content Filtering.} PII and harmful content recognition and filtering is essential for reducing the model's toxicity. 
    Recent works employ heuristic rule-based filtering strategies. 
    However, these methods require rule curation manually and suffer from poor performance on scalability, precision, and recall, making them unsuitable for building robust, and automated data processing pipelines. 
    We adopt a combined approach of heuristic rules and model-based methods.
    Specifically, the former primarily includes fixed site templates, URL blacklists, regular expressions, word count and repeated n-gram character ratios, while the latter follows a iterative distillation method.
    We first obtain a set of harmful data by GPT-4 and human annotation as seeds to train a FastText~\cite{fasttext-classfication} model for classification, marking the PII, advertisement, copyright statements, and meaningless special characters as negative samples.
    Hard samples are mined to iterative improve the previous model and results in a generalized harmful content classification model ultimately. 
    The probability that data points classified as negative is taken as a proxy of quality scores for PII an harmful content filtering.
    Additionally, data with low language scores is filtered, mostly meaningless texts.
    
\end{itemize}

\begin{table}[htbp]
    \centering
    \small
    \caption{Loss of different deduplication strategies for 2B foundation models on held-out set.}
    \setlength\tabcolsep{0.8mm}{
    \begin{tabular}{lcccc}
    \toprule
    & \textbf{Loss (ZH)} & \textbf{Loss (EN)} \\ \midrule
     w/o dedup & 3.136 & 2.893 \\
     w/ document-level dedup & 3.083 & 2.877 \\ 
     w/ multi-level dedup & \textbf{3.006} & \textbf{2.874} \\
    \bottomrule
    \end{tabular}}
    \label{tab:dedup}
\end{table}

\begin{figure}
    \centering
    \includegraphics[width=\linewidth]{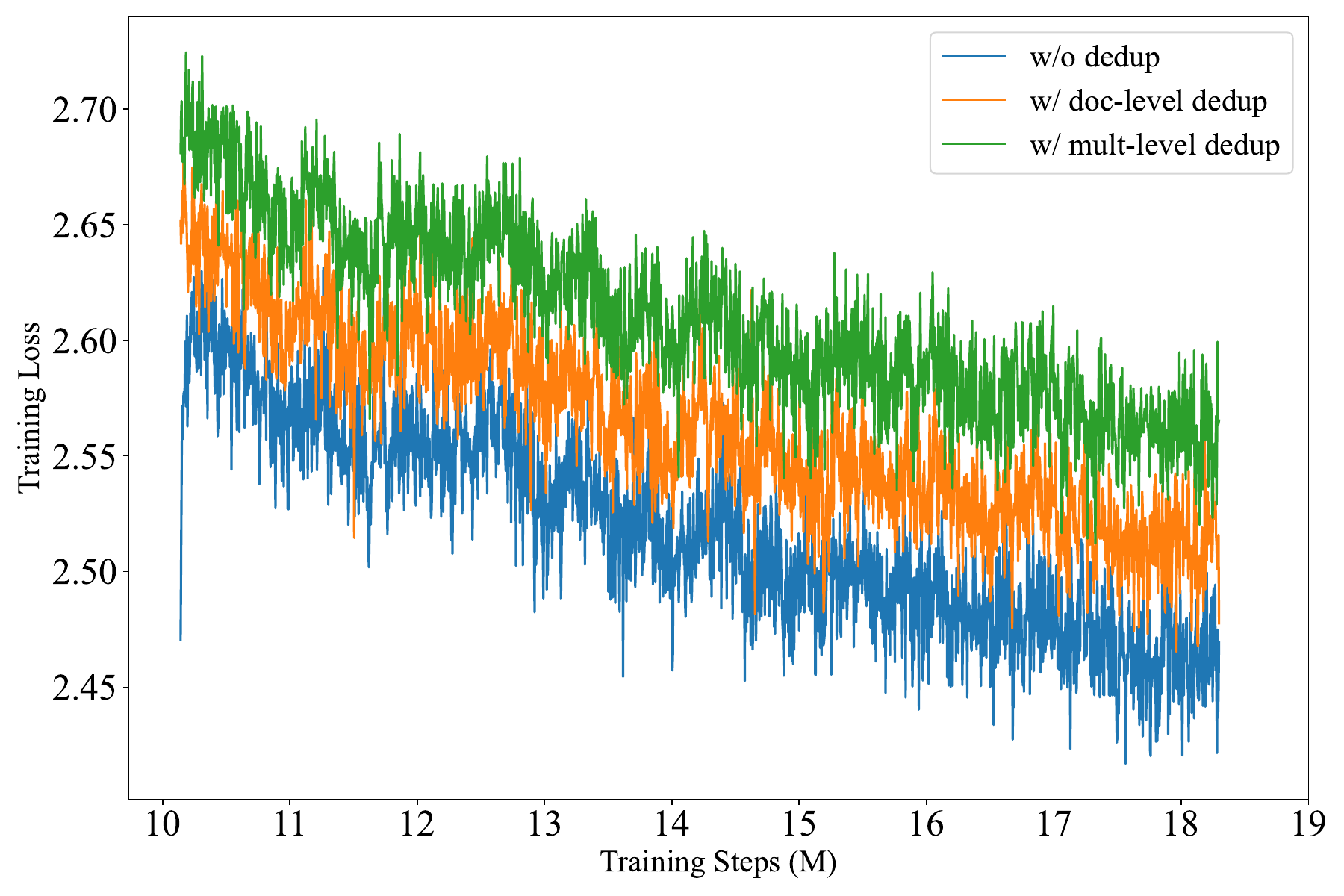}
    \caption{The training loss with respect to different deduplication strategies on 2B models.}
    \label{fig:dedup-training-loss}
\end{figure}

We conduct ablation studies with different deduplication strategies by continued pre-training several 2B models with the same architecture as BaichuanSEED.
The result can be found in Table~\ref{tab:dedup}.
With the same training tokens, our global multi-granularity deduplication strategy achieves the lowest loss compared with the strategy ``without deduplication'' on both Chinese and English test set by -4.15\% and -0.66\% respectively, improving the model compression ability significantly~\cite{compression-iclr2024}.
Notably, our global multi-granularity deduplication strategy increases the training loss, as shown in Figure~\ref{fig:dedup-training-loss}.
We argue that our deduplication strategy leads to better diversity, making the prediction of the foundation model harder in the pre-training stage.

\begin{table}[htbp]
    \centering
    \small
    \caption{The data mixture of our pre-training data.}
    \setlength\tabcolsep{0.8mm}{
    \begin{tabular}{lccc}
    \toprule
    \textbf{Domain} & \textbf{\#Tokens} & \textbf{\#Token/Doc} & \textbf{Weight} \\
     \midrule
     Web / Chinese & 405B & 361 & 13.5\%\\
     Web / Engligh & 945B & 752 & 31.5\% \\
     KID / Chinese & 300B & 10,362 & 10\% \\
     KID / English & 450B & 16,172 & 15\% \\
     Code & 750B & 3,157 & 25\% \\
     Others & 150B & 409 & 5\% \\ \midrule
     \textbf{Pre-training Data} & 3000B & 4,558 & 100\%\\ 
    \bottomrule
    \end{tabular}}
    \label{tab:data-mixture}
\end{table}

\noindent\textbf{Data Mixture.}\quad 
We conduct several empirical experiments to determine the data mixture of each domain, with some details that can be found in Section~\ref{sec:effect-of-hkd}.
We then train our BaichuanSEED with the fixed data mixture from scratch as shown in Table~\ref{tab:data-mixture}.

\subsubsection{Other Principles}
\label{sec:other-principles}

\noindent\textbf{Synthetic Data.}\quad 
Synthetic data is widely used in SFT to enhance the difficulty, complexity, and diversity of constructed instructions~\cite{self-instruct, evol-instruct}.
However, some studies demonstrate that while synthetic data can improve model performance on coding and mathematics benchmarks, it may lead to a loss of instruction following ability after SFT~\cite{chenjie-synthetic-pretraining}, making the impact of synthetic data unclear and introducing additional confounding variables. 
Our technical report focuses on training a completely pure baseline to explore the limits of model capability influenced by data collection and deduplication. 
Therefore, we did not incorporate synthetic data when training BaichuanSEED. 
We plan to explore the impact of synthetic data in future work.

\noindent\textbf{Fine-grained Curation of Data.}\quad 
We directly prompt GPT-4 to obtain a set of knowledge category labels of training data and train a model to classify the full dataset, enhancing our fine-grained understanding. 
These labels are used to further explore the potential on mathematics of our model, as discussed in Section~\ref{sec:potential-on-math}. 
They can also be applied to more detailed curriculum learning exploration.
Additionally, we make effort to ensure that the evaluation datasets are excluded from the training data of BaichuanSEED.

\subsection{Training Details}

We train on $3$T tokens with a sequence length of $16$K and a global batch size of $224$.
A cosine learning rate scheduler is used, with an initial learning rate of $3e\mathrm{-}4$ and a final learning rate of $3e\mathrm{-}5$. 

\section{Supervised Fine-tuning}

In this section, we focus on the details of the stage of SFT, including our meticulously designed instruction construction and our training settings.

\subsection{SFT Data}

During the stage of SFT, we utilize a self-constructed dataset for training. 
This comprehensive dataset includes mathematics, logical reasoning, coding, creative writing, brainstorming, and multi-turn dialogues, with approximately $450$K samples. 
Based on prior research~\cite{COIG-CQIA-Quality-is-all-you-need,llama2}, we make optimizations to the SFT data, primarily including clustering massive data to enhance the diversity, synthesizing the instructions to increase the complexity.
Additionally, we employ human annotations for part of data to improve the quality.

\subsection{Training Details}

We use a constant learning rate schedule with an initial learning rate of $2e\mathrm{-}5$, a batch size of $40$, and a sequence length of $16$K tokens.
Each sample includes three parts: system, which is optimal, prompt, and answer. 
To enhance training efficiency and ensure sequences were filled, we employ sample packing for data concatenation and utilize Flash Attention 2~\cite{flashattention-2} to accelerate the training.

Ultimately, we selected a 6-epoch checkpoint for validation. 
This is more than the usual number of epochs for SFT training, comparing to Llama2, which used only 2 epochs.
Empirically, more epochs yields better results for small size models, especially when the amount of pre-training token is insufficient.
This might not be the optimal solution, as the ideal number of epochs can vary with different checkpoints. 
Nevertheless, this does not affect our evaluation and comparison of the alignment performance of the foundation models.
We discuss the SFT evaluation results in Section~\ref{scaling-curve}.

\section{Evaluation}

In this section, we first explore the scaling law of BaichuanSEED's downstream task performance with respect to the amount of training tokens. 
We elucidate the consistency and predictability during the training of BaichuanSEED. 
Furthermore, we evaluate our model and a series of 7B LLM baselines on several comprehensive benchmarks.
We also assess the model's generalization ability across a series of carefully selected downstream task benchmarks in coding, mathematics, common-sense reasoning, and reading comprehension, demonstrating the robustness of our model.

\subsection{Scaling Curves}
\label{scaling-curve}

Since we emphasize that appropriate data collection and deduplication can lead to comparable performance, which requires significantly less manual effort and computational cost, compared to meticulous data selection during the pre-training phase. 
However, the absence of data selection may result in inevitable noise.
The process might raise concerns about whether training on a larger yet potentially lower-quality dataset can truly yield benefits or even have a negative impact such as drastic training fluctuation, and worse downstream performance.
Consequently, we demonstrate the robustness of BaichuanSEED from two dimensions: \textbf{Consistency} and \textbf{Predictability}.

\begin{figure}
    \centering
    \includegraphics[width=\linewidth]{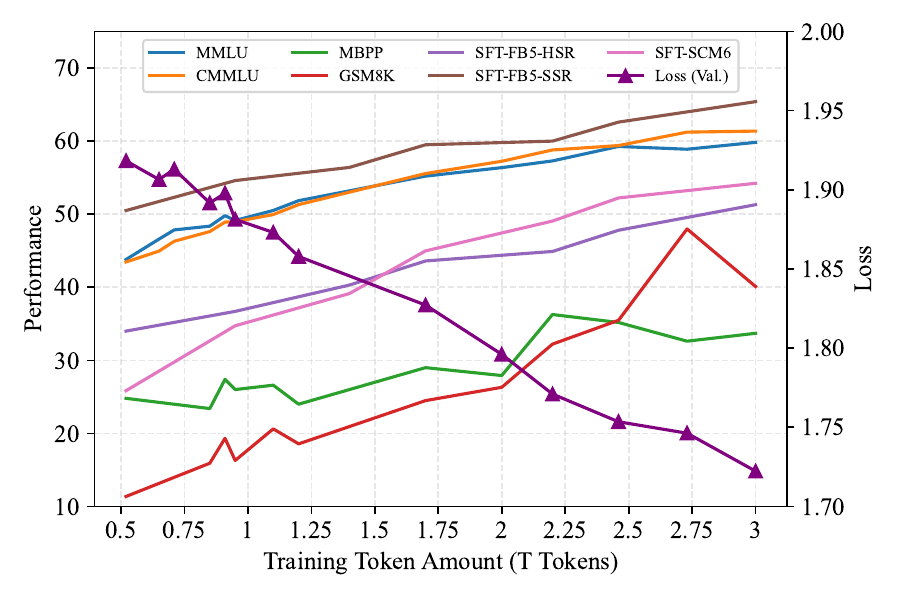}
    \caption{The performance scaling curve of BaichuanSEED with respect to the amount of training tokens.}
    \label{fig:consistency}
\end{figure}

\noindent\textbf{Consistency.}\quad  
We define consistency as the consistent trend of model capability growth with the number of training tokens before and after SFT. 
Specifically, model capability is reflected by metrics including the loss on test sets, comprehensive benchmark performance of base models, and the SFT benchmark performance of SFT models.
We illustrate trends of the metrics with respect to the amount of training tokens during pre-training, as shown in Figure~\ref{fig:consistency}.
The comprehensive benchmarks consist of MMLU~\cite{MMLU}, CMMLU~\cite{CMMLU}, MBPP~\cite{MBPP}, GSM8K~\cite{GSM8K}.
While the SFT benchmarks encompass FollowBench~\cite{followbench} and SuperCLUE-Math6~\cite{superclue-math6}.
The SFT benchmark results are evaluated on different checkpoints of the foundation model after the same SFT process.
BaichuanSEED exhibits remarkable consistency across all evaluation benchmarks, indicating the model's generalization capability. 
It indicates that BaichuanSEED has not been specialized to optimize for a specific task or evaluation benchmark, retaining the potential to transfer to specific downstream tasks.

Notably, some models may incorporate excessive synthetic data during the pre-training phase, creating a bubble impression of strong generalization capabilities and hindering the potential to improve during the SFT phase, even affecting the instruction-following capability~\cite{chenjie-synthetic-pretraining}. 
We argue that pre-training and SFT are both indispensable, and excessively exploiting SFT may hinder the community from being aware of the foundational models' capabilities.
In contrast, our model exhibits consistent gains across both pre-training and SFT evaluation benchmarks, validating the decision to minimize using synthetic data during the pre-training phase.
Additionally, the consistently increasing trend on several benchmarks validates the stability of our training strategy.
Stability is highly essential in LLM development, where many approaches often sacrifice training accuracy to ensure training stability~\cite{opt,glm-130b}.


\begin{figure}
    \centering
    \includegraphics[width=\linewidth]{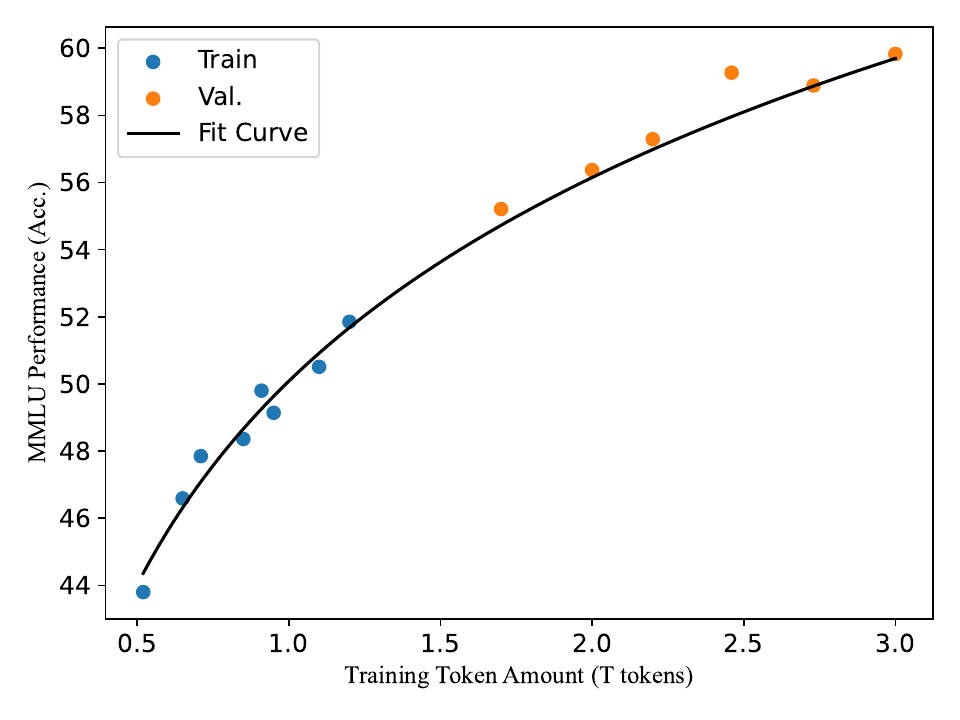}
    \caption{The curve fit by the first half performance of BaichuanSEED with respect to the amount of training tokens on MMLU.}
    \label{fig:predictability}
\end{figure}

\noindent\textbf{Predictability.}\quad 
We define predictability as the ability to forecast the model's capabilities at later checkpoints based on its performance during the early stages of training.
Predictability is especially crucial for developers, as having a forward-looking understanding of the model's performance trends during the early stages of training can facilitate rapid iteration and minimize unnecessary resource costs.
Taking BaichuanSEED's performance on MMLU as an example, we employ a logarithmic curve to fit the performance on MMLU with respect to the amount of training tokens, allowing us to predict the potential of our model. 
Specifically, we take points from the first half for extrapolation and fit for the points in the latter half perfectly, demonstrating the predictability ability of our model, as shown in Figure~\ref{fig:predictability}. 


\subsection{Comprehensive Benchmarks}

\begin{table*}[t]
    \centering
    \small
    \caption{The performance on comprehensive benchmarks.}
    \setlength\tabcolsep{1mm}{
    \begin{tabular}{lccccccccc}
    \toprule
    & \textbf{Training Tokens} & \textbf{MMLU} & \textbf{CMMLU} & \textbf{AGIEval} & \textbf{C-Eval} & \textbf{MMLU-Pro} & \textbf{LiveBench}\\
    & & (5-shot) & (5-shot) & (0-shot) & (5-shot) & (5-shot) & (0-shot)\\
    \midrule
    Baichuan2-7B & 2.6T & 54.65 & 56.95 & 28.95 & 56.19 & 21.65 & -\\
    Baichuan2-13B & 2.6T & 59.83 & 61.32 & 24.07 & 58.10 & 26.59 & -\\
    Qwen1.5-7B & 3T & \underline{62.19} & \textbf{71.84} & \textbf{39.46} & \textbf{73.64} & \underline{30.30} & -\\
    Llama3-8B & 15T & \textbf{66.57} & 50.68 & 26.74 & 49.89 & \textbf{35.30} & - \\ 
    OLMo-7B & 2.5T & 28.40 & 25.55 & 19.89 & 27.27 & 13.05 & - \\ 
    MAP-Neo-7B & 4.5T & 58.18 & 55.06 & \underline{33.87} & 57.50 & 26.89 & - \\ 
    \midrule
    BaichuanSEED & 3T & 60.25 & \underline{62.09} & 31.07 & \underline{61.58} & 26.57 & -\\ 
    \midrule \midrule
    Baichuan2-7B-Chat & 2.6T & 54.35 & 55.36 & 35.29 & 55.09 & 25.11 & 12.89\\
    Baichuan2-13B-Chat & 2.6T & 57.28 & \underline{61.32} & 30.15 & 58.04 & 28.03 & 13.04 \\
    Qwen1.5-7B-Chat & 3T & \underline{61.49} & \textbf{68.02} & \textbf{39.29} & \textbf{68.96} & 16.29 & 16.78\\
    Llama3-8B-Instruct & 15T & \textbf{67.10} & 51.66 & \underline{38.37} & 50.71 & \textbf{41.88} & \textbf{25.91}\\
    OLMo-7B-SFT & 2.5T & 47.49 & 35.49 & 29.12 & 35.43 & 17.99 & 8.80  \\
    MAP-Neo-7B-SFT & 4.5T & 58.31 & 55.24 & 37.98 & 55.58 & \underline{30.24} & 14.35 \\
    \midrule
    BaichuanSEED-SFT & 3T & 60.15 & 60.84 & 32.62 & \underline{59.41} & 29.63 & \underline{18.32} \\
    
    \bottomrule
    \end{tabular}}
    \label{tab:comprehensive}
\end{table*}

We present the performance of BaichuanSEED and a series of 7B models as baselines across a comprehensive set of evaluation benchmarks.

\noindent\textbf{Baselines.}\quad 
The selected baselines are all cutting-edge open-sourced LLMs, including the base and chat or instruct version of Llama family~\cite{llama3}, Baichuan series~\cite{baichuan}, Qwen series~\cite{qwen2}, MAP-Neo~\cite{map-neo}, and OLMo~\cite{olmo}. 

\noindent\textbf{Benchmarks.}\quad 
The benchmarks encompass MMLU, CMMLU, AGIEval~\cite{zhong2023agieval}, C-Eval, MMLU-Pro~\cite{MMLU-Pro}, and LiveBench~\cite{white2024livebench}.
We also carefully select representative benchmarks for downstream task performance evaluation, including MBPP, HumanEval~\cite{HumanEVAL} for coding, GSM8K, MATH~\cite{MATH-dataset} for mathematics, TriviaQA~\cite{2017arXivtriviaqa} for commonsense reasoning, and HellaSwag~\cite{zellers2019hellaswag} for reading comprehension.
These benchmarks typically cover a wide range of subjects, prompting the LLMs to directly answer questions, which may be in the form of single-choice or short-answer. 
We take accuracy as the evaluation metric.

\begin{table*}[t]
    \centering
    \small
    \caption{The performance on code, mathematics, commonsense reasoning, and reading comprehension benchmarks.}
    \setlength\tabcolsep{1mm}{
    \begin{tabular}{lcccccccccc}
    \toprule
    & \textbf{Training Tokens} & \textbf{MBPP} & \textbf{HumanEval} & \textbf{MATH} & \textbf{GSM8K} & \textbf{TriviaQA} & \textbf{HellaSwag}\\ 
    & & (3-shot) & (0-shot) & (4-shot) & (4-shot) & (0-shot) & (0-shot)\\ \midrule
    Baichuan2-7B & 2.6T & 25.40 & 17.68 & 5.94 & 25.02 & 53.73 & 67.56 \\
    Baichuan2-13B & 2.6T & 30.88 & 17.07 & 10.68 & 52.08 & \underline{58.73} & 71.09\\
    Qwen1.5-7B & 3T & \underline{36.60} & \textbf{53.05} & \textbf{21.08} & \textbf{54.74} & 50.92 & \underline{72.64} \\
    Llama3-8B & 15T & \textbf{44.60} & \underline{26.22} & 13.44 & 50.11 & \textbf{65.23} & \textbf{74.54} \\ 
    OLMo-7B & 2.5T & 21.00 & 11.59 & 1.72 & 2.00 & 49.81 & 70.31 \\ 
    MAP-Neo-7B & 4.5T & 25.90 & 7.93 & \underline{15.14} & \underline{53.90} & 54.80 & 67.85 \\ 
    \midrule
    BaichuanSEED & 3T & 34.12 & 21.34 & 9.84 & 38.81 & 45.92 & 70.20 \\ \midrule \midrule
    Baichuan2-7B-Chat & 2.6T & 22.40 & 15.24 & 8.70 & 32.37 & 44.65 & 69.18 \\
    Baichuan2-13B-Chat & 2.6T & 26.30 & 18.90 & 8.62 & 56.79 & 53.47 & 72.32 \\
    Qwen1.5-7B-Chat & 3T & 12.58 & \textbf{29.27} & 13.12 & 56.10 & 10.22 & 72.81 \\    
    Llama3-8B-Instruct & 15T & \textbf{52.17} & 21.34 & \underline{25.62} & \textbf{78.17} & \textbf{63.37} & 71.45 \\
    OLMo-7B-SFT & 2.5T & 25.16 & 19.51 & 2.52 & 17.66 & 42.87 & \underline{72.62} \\
    MAP-Neo-7B-SFT & 4.5T & 33.66 & \textbf{29.27} & \textbf{30.86} & \underline{70.28} & \underline{53.82} & 68.48 \\
    \midrule
    BaichuanSEED-SFT & 3T & \underline{37.60} & \underline{23.17} & 14.06 & 53.98 & 43.92 & \textbf{73.03} \\
    
    \bottomrule
    \end{tabular}}
    \label{tab:coding-math}
\end{table*}

\noindent\textbf{Overall Performance.}\quad 
The overall performance is shown in Table~\ref{tab:comprehensive}, while the details across various subjects can be found in Appendix~\ref{Detail-eval-result}. 
Compared to the Llama family models, BaichuanSEED shows significant gains in Chinese evaluation benchmarks, such as CMMLU and C-Eval. 
We suppose that the incorporation of knowledge intensive Chinese data, such as high-quality textbooks and academic papers, leading to the improvement.
Moreover, our foundation model surpasses Baichuan2-7B and Baichuan2-13B in comprehensive English benchmarks such as MMLU~(+10.2\%, +0.7\%), AGIEval~(+7.3\%, +29.1\%), and MMLU-Pro~(+22.7\%, -0.1\%), and is comparable to Qwen1.5-7B and MAP-Neo. 
Although our model still lags behind Llama3-8B, we argue that training with only one-fifth of Llama's tokens may contribute to the result, which suggests the potential for comparable or even superior performance at the same scale. 
LiveBench, a benchmark with high timeliness, effectively assesses the model's generalization ability, preventing potential benchmark leakage. 
Our model also outperforms Baichuan2-7B-Chat~(+42.1\%), Qwen1.5-7B-Chat~(+9.2\%), OLMo-7B-SFT~(+108.2\%), and MAP-Neo-7B-SFT~(+27.7\%) on LiveBench, likely due to its focus on acquiring extensive world knowledge and language patterns, not limited to overfit on specific optimization tasks.
To conclude, our BaichuanSEED achieve competitive performance without any elaborate data selection and benchmark optimazation, even compared to the cutting-edge commercial LLMs.

\noindent\textbf{Downstream task Performance.}\quad 
LLMs are typically expected to exhibit strong generalization capabilities across multiple downstream tasks. 
The detained performance is shown in Table~\ref{tab:coding-math}.
Our fine-tuned model BaichuanSEED-SFT consistently achieve second best in code tasks, due to the rich logical reasoning found in high-quality data. 
However, our model underperforms in mathematics, trailing Llama3 and MAP-Neo by nearly 10 points on MATH and 25 points on GSM8K, respectively.
For example, MAPNeo achieved only 5\% accuracy on MATH and 21\% on GSM8K before annealing, while significant improvement after annealing with high ratio of mathematical data~\cite{map-neo}.
To emphasize, BaichuanSEED tries to be a completely pure model, without training on synthetic data or up-sampling any downstream task data in pre-training, and utilizing general instructions during SFT. 
Therefore, our foundation model still has significant untapped potential, especially in code and mathematical tasks, discussed in Section~\ref{sec:potential-on-math}. 
This can be achieved through further pre-training with an increased proportion of downstream task data~\cite{deepseekmath,deepseek-coder-v2} or high-quality data annealing training~\cite{MiniCPM,llama3-herd}.
Our foundation model is comparable to Qwen1.5 and Llama on HellaSwag, and performs best after fine-tuning.
Notably, our model shows no significant changes after alignment, except for a slight improvement in math benchmarks, attributed to the enhanced instruction following and comprehension abilities gained during SFT. 
In contrast, Qwen1.5 experienced a decline in MMLU-Pro and downstream code and math tasks, highlighting the importance of a completely pure training process to ensure the consistency after alignment.


\section{Discussion}

In this section, we conduct a series of experiments to explore the effectiveness of strategies employed in the training of BaichuanSEED. 
Firstly, thanks to the precise categorization of subjects as mentioned in Section~\ref{sec:other-principles} to adjust the proportion of mathematics, we validate the potential of our model in mathematical tasks under the continued pre-training strategy. 
Secondly, we investigate the optimal ratio of high-density knowledge data during the pre-training phase using both cold start and continued pre-training methods on a 2B same-architecture model. 




\begin{figure*}
    \centering
    \includegraphics[width=\linewidth]{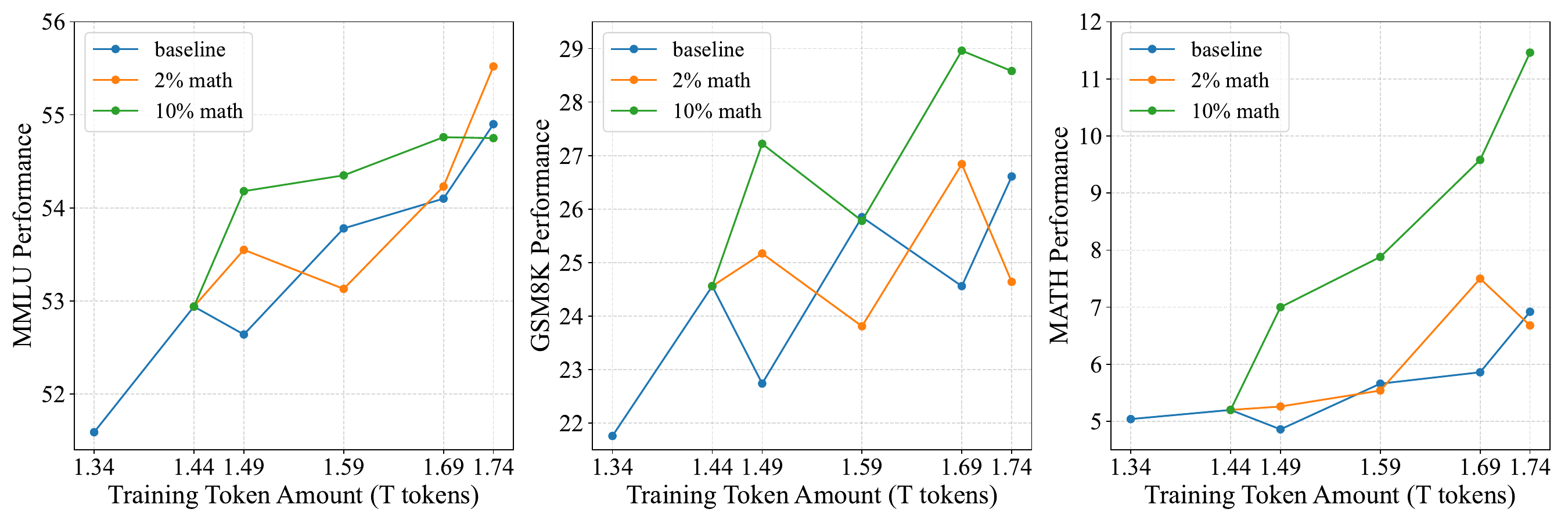}
    \caption{The performance of different checkpoints continued pre-training from a 1.44T token BaichuanSEED checkpoint with different proportion of mathematics-related data evaluated on MMLU, GSM8K, and MATH.}
    \label{fig:math-potential}
\end{figure*}

\subsection{Effect of Knowledge Intensive Data}
\label{sec:effect-of-hkd}

Knowledge Intensive Data refers to data that contains massive knowledge per token. 
We argue that LLMs trained on KID can achieve better performance in the same training steps~\cite{MiniCPM,Phi1-5}. 
We extensively collect public Chinese and English academic papers and books as stated in Section~\ref{data-collection}. 
However, the introduction of KID inevitably down-sampling others, which may hinder the model from possessing comprehensive world knowledge or language patterns. 
Therefore, determining the appropriate proportion of KID is crucial.

We explore the optimal proportion of KID in two settings: training from scratch and continued pre-training. 
Considering the computational cost, we conduct experiments on a 2B model with the same architecture as BaichuanSEED, training from scratch or continued pre-training 100B tokens from a model that trained on 1.5T tokens and have already obtained emergent abilities, for two settings respectively. 
We search for the sampling proportion of KID from 0\% to 50\%, while keeping the remaining distribution of others consistent with Table~\ref{tab:data-mixture}.
We evaluate the language modeling capabilities on a private, high-quality corpus that constructed by a wide range of domains such as encyclopedias, web pages, papers, and news. 

\begin{table}[htbp]
    \centering
    \small
    \caption{The performance of models trained with different proportion of knowledge intensive data.}
    \setlength\tabcolsep{0.8mm}{
    \begin{tabular}{lcccccc}
    \toprule
     & \multicolumn{3}{c}{\textbf{Train from Scratch}} & \multicolumn{3}{c}{\textbf{Continued Pretrain}} \\
    \cmidrule(r){2-4}\cmidrule(l){5-7}
     KID Prop. & 0\% & 25\% & 50\% & 0\% & 25\% & 50\% \\
     \midrule
     PPL & 14.24  & \textbf{13.68}  & 13.86  & 12.21  & \textbf{11.73}  & 11.80 \\
    \bottomrule
    \end{tabular}}
    \label{tab:hkd}
\end{table}

The results are shown in Table~\ref{tab:hkd}. 
The model trained with 25\% KID achieved the lowest perplexity, i.e. best language modeling capabilities. 
This suggests that the inclusion of KID accelerates the model's acquisition of language modeling capabilities.
Consequently, we adopted 25\% as the final proportion of knowledge intensive data for training BaichuanSEED based on the scaling law.

\subsection{Potential on Mathematics}
\label{sec:potential-on-math}

Taking mathematics reasoning as an example, we explore the potential of BaichuanSEED on a wide range of downstream tasks.
Specifically, we first employ a classification model to perform a finer categorization of high-density knowledge data on subjects to better utilize this data. 
We then conduct continued pre-training experiments by altering the data proportions to verify the potential of our model. 
The experimental results indicate that a higher proportion of mathematics data significantly enhances the model's performance in mathematical downstream tasks, such as GSM8K and MATH, without compromising the comprehensive evaluation benchmarks.

To address the details, we increase the proportion of mathematics-related data to 2\% and 10\%, conducting 300B token continued pre-training on a checkpoint with 1.44T tokens of BaichuanSEED, as illustrated in Figure~\ref{fig:math-potential}. 
At the 2\% proportion, we do not observe significant gains on the mathematics benchmarks. 
However, when we aggressively raise the proportion to 10\%, notable improvements are evident. 
To be specific, the 1.74T checkpoint shows a 7.4\% and 65.6\% increase on GSM8K and MATH respectively, with only a 0.3\% fluctuation on MMLU, compared to the baseline. 
Therefore, we argue that knowledge intensive data has substantial potential, especially when enhancing a language model's mathematical abilities. 
By searching optimal data proportions and employing curriculum learning strategies, the value of these data can be further captured.
In addition, it validates the immense potential of our model in downstream tasks.


\section{Related Works}
\label{sec:related-works}


Several institutions have published detailed technical reports and open-sourced the weights of their cutting-edge models, to promote the development of the LLM field. 
Recently, numerous models made efforts on meticulous data filtering~\cite{llama3, falcon-refinedweb, wrap, dolma}, innovative architectural exploration~\cite{gqa, mixtral-8-7b-moe,mamba}, and optimization for vertical downstream tasks~\cite{deepseek-coder-v2, deepseekmath}. 
As a baseline model, BaichuanSEED focus on the exploration of the limits of model capabilities through data collection and deduplication, avoiding any specific optimization. 
Nevertheless, our model achieves competitive performance on comprehensive benchmarks, comparable to many open-source commercial models of similar scales. 
These representative models include earlier versions of Baichuan~\cite{baichuan}, the Llama family~\cite{llama2,llama3-herd}, the Qwen series~\cite{qwen2}, the Deepseek series~\cite{deepseek-moe,deepseek-v2}, and the Phi series~\cite{Phi1-5,phi-3}.
Moreover, there are several initiatives dedicated to disclosing technical details, such as training datasets and code, including MAPNeo~\cite{map-neo}, OLMo~\cite{olmo}, Pythia~\cite{pythia}, and Amber~\cite{llm360-amber}.


\section{Conclusions}

This technical report introduces BaichuanSEED, a pure yet competitive 7 billion parameter large language model trained from scratch to explore its limits without elaborate data selection and specific downstream task optimization. 
We pre-train the model on 3 trillion tokens of high-quality data and supervised fine-tune it with a straightforward yet effective process. 
Guided by the pretraining principles that collection to scale up and reweighting to improve data quality, our model achieves comparable performance with leading commercial models on several comprehensive benchmarks. 
We also conduct heuristic experiments to explore the model’s potential for further optimization in downstream tasks. 
All training details are publicly available, with the hope that our technical report would benefit the community to further unveil the skyline of data on large language models.



\newpage


\bibliography{main}

\newpage

\balance
\appendix


\begin{figure*}
    \centering
    \includegraphics[width=\linewidth]{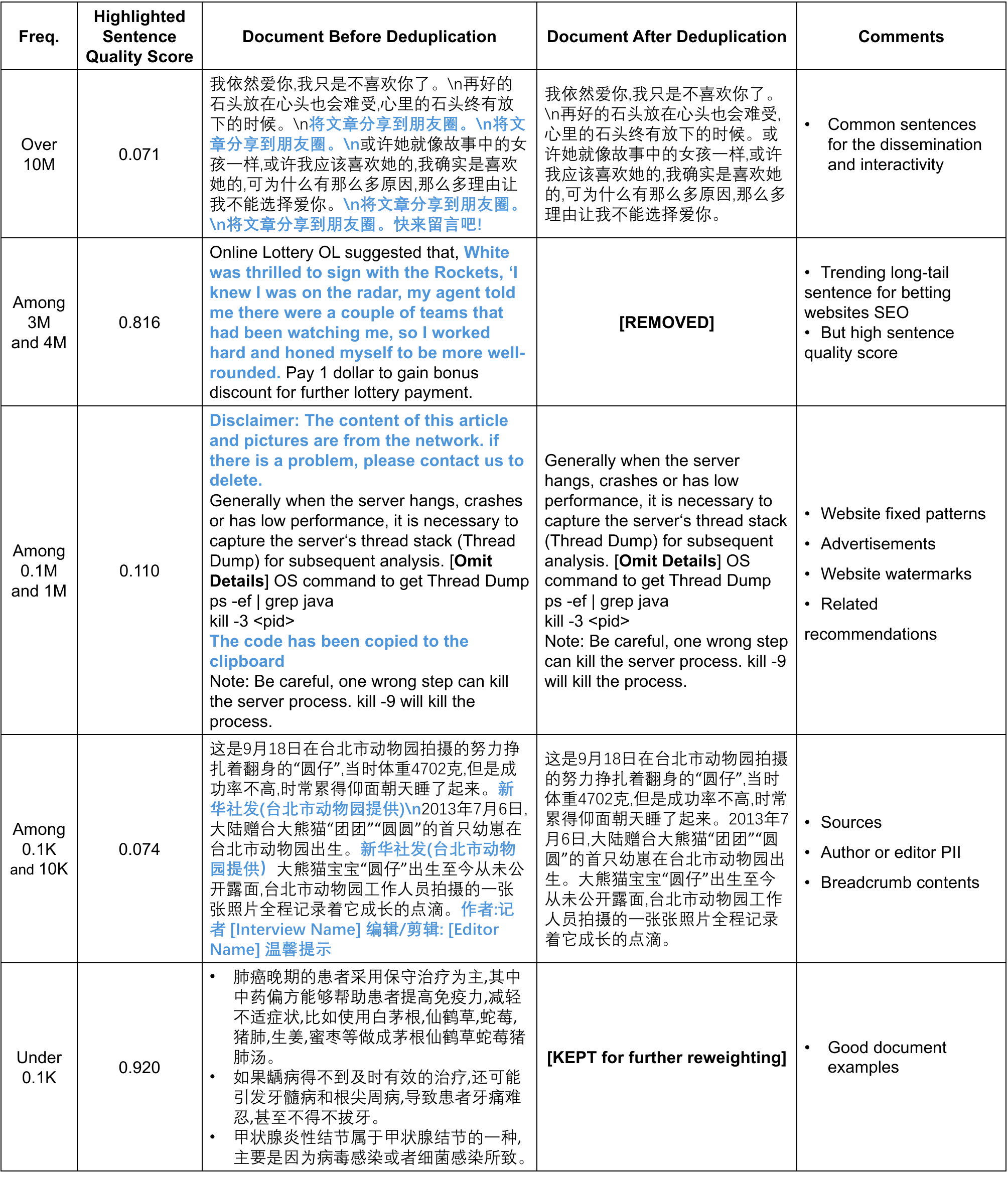}
    \caption{The cases with different frequency before and after our global multi-granularity deduplication. We provide the frequency, the quality score of highlighted sentences, and our comments on each case. The contents highlighted in blue are removed after deduplication. }
    \label{fig:dedup-cases}
\end{figure*}

\begin{table*}[htbp]
    \centering
    \small
    \caption{The detailed performance on MMLU, CMMLU, and AGIEval.}
    \resizebox{\textwidth}{!}{
    \setlength\tabcolsep{1mm}{
    \begin{tabular}{lccccccccccccccc}
    \toprule
    \textbf{Models} & \multicolumn{5}{c}{\textbf{MMLU}} & \multicolumn{6}{c}{\textbf{CMMLU}} & \multicolumn{4}{c}{\textbf{AGIEval}}\\
    \cmidrule(r){2-6}\cmidrule(l){7-12}\cmidrule(l){13-16}
    & \textbf{Avg.} & \textbf{HUM} & \textbf{SOC} & \textbf{STEM} & \textbf{OTH} & \textbf{Avg.} & \textbf{CHI} & \textbf{SOC} & \textbf{STEM} & \textbf{HUM}  & \textbf{OTH}  & \textbf{Avg.}  & \textbf{ZH}  & \textbf{EN}  & \textbf{Gaokao}  \\ \midrule 
    Baichuan2-7B & 54.65 & 60.17 & 62.25 & 45.03 & 56.31 & 56.95 & 59.31 & 62.10 & 43.42 & 61.05 & 61.19 & 28.95 & 31.17 & 25.99 & 35.24 \\ 
    Baichuan2-13B & 59.83 & 65.56 & 68.54 & 48.54 & 62.55 & 61.32 & 62.59 & 65.93 & 47.36 & 66.79 & 65.63 & 24.07 & 35.96 & 8.21 & 39.29 \\ 
    Llama3-8B & \textbf{66.57} & \textbf{70.38}& \textbf{75.93} & 56.33 & \textbf{68.97} & 50.68 & 44.78 & 52.67 & 43.54 & 52.96 & 53.88 & 26.74 & 31.45 & 20.46 & 34.81 \\ 
    Qwen1.5-7B & 62.19 & 65.73 & 70.02 & 53.49 & 64.15 & \textbf{71.84} & \textbf{73.01} & \textbf{74.09} & \textbf{63.07} & \textbf{76.01} & \textbf{74.84} & \textbf{39.46} & \textbf{53.08} & 21.31 & \textbf{59.26} \\ 
    OLMo-7B & 28.40 & 25.13 & 30.01 & 29.99 & 28.52 & 25.55 & 24.97 & 25.27 & 25.27 & 26.58 & 25.36 & 19.89 & 18.84 & 21.29 & 19.44 \\
    MAP-Neo-7B & 58.18 & 60.88 & 66.83 & \textbf{59.86} & 49.73 & 55.06 & 55.88 & 59.44 & 45.58 & 62.14 & 53.24 & 33.87 & 33.91 & \textbf{33.83} & 37.92 \\
    \midrule
    BaichuanSEED & 60.25 & 63.78 & 68.85 & 50.22 & 61.39 & 62.09 & 61.36 & 66.20 & 47.57 & 66.13 & 65.89 & 31.07 & 34.70 & 21.17 & 40.28 \\ 
    \midrule \midrule
    Baichuan2-7B-Chat & 54.35 & 60.27 & 62.37 & 43.48 & 57.10 & 55.36 & 56.24 & 58.75 & 43.61 & 61.05 & 61.19 & 35.29 & 37.98 & 21.71 & 42.70 \\ 
    Baichuan2-13B-Chat & 57.28 & 62.58 & 65.65 & 46.72 & 59.67 & 59.74 & 61.10 & 64.54 & 46.37 & 65.10 & 63.19 & 30.15 & \textbf{41.02} & 15.65 & 46.43 \\ 
    Llama3-8B-Instruct & \textbf{67.10} & \textbf{70.75} & \textbf{76.88} & 57.08 & \textbf{69.29} & 51.66 & 45.84 & 52.92 & 44.21 & 52.49 & 57.53 & 38.37 & 38.30 & 38.46 & 41.53\\
    Qwen1.5-7B-Chat & 61.49 & 66.59 & 69.12 & 52.65 & 62.26 & \textbf{68.02} & \textbf{68.58} & \textbf{71.21} & \textbf{59.31} & \textbf{71.83} & \textbf{69.92} & \textbf{39.29} & 39.87 & \textbf{38.52} & 46.26 \\ 
    OLMo-7B-SFT & 47.49 & 48.48 & 55.78 & 51.23 & 39.02 & 35.49 & 33.60 & 37.81 & 30.81 & 33.84 & 38.82 & 29.12 & 27.37 & 31.46 & 30.42 \\
    MAP-Neo-7B-SFT & 58.31 & 60.44 & 66.22 & \textbf{58.23} & 51.90 & 55.24 & 55.73 & 57.51 & 47.25 & 63.43 & 53.86 & 37.98 & 38.22 & 37.66 & 43.69 \\
    \midrule
    BaichuanSEED-SFT & 60.15 & 65.10 & 67.82 & 49.68 & 62.70 & 60.84 & 60.28 & 64.76 & 49.07 & 64.36 & 65.46 & 32.62 & 40.28 & 20.42 & \textbf{47.44} \\ 
    \bottomrule
    \end{tabular}}
    }
    \label{tab:mmlu}
\end{table*}

\section{Detailed Evaluation Results}
\label{Detail-eval-result}

\begin{table*}[htbp]
    \centering
    \small
    \caption{The detailed performance on MMLU-Pro.}
    \resizebox{\textwidth}{!}{
    \setlength\tabcolsep{1mm}{
    \begin{tabular}{lccccccccccccccc}
    \toprule
    \textbf{Models}  & \textbf{Avg.} & \textbf{BIO} & \textbf{BUS} & \textbf{CHEM} & \textbf{CS} & \textbf{ECO} & \textbf{ENG} & \textbf{HEA} & \textbf{HIST}  & \textbf{LAW}  & \textbf{MATH}  & \textbf{OTH}  & \textbf{PHIL}  & \textbf{PHY}  & \textbf{PSY} \\ \midrule 
    Baichuan2-7B & 21.65 & 33.61 & 19.90 & 14.31 & 21.95 & 27.01 & 12.80 & 24.33 & 23.10 & 16.17 & 14.80 & 25.22 & 22.65 & 15.09 & 32.21 \\
    Baichuan2-13B & 26.59 & 41.84 & 26.36 & 15.28 & 27.32 & 33.77 & 16.10 & 30.93 & 27.82 & 15.53 & 19.62 & 31.28 & 29.46 & 19.71 & 37.22 \\
    Llama3-8B & \textbf{35.30} & \textbf{56.40} & \textbf{32.07} & \textbf{24.82} & \textbf{33.66} & \textbf{46.68} & \textbf{25.49} & \textbf{43.28} & \textbf{36.22} & \textbf{19.52} & \textbf{30.42} & \textbf{40.48} & \textbf{31.41} & \textbf{53.26} & 41.45 \\
    Qwen1.5-7B & 30.30 & 43.51 & 28.01 & 23.59 & 32.93 & 39.57 & 21.98 & 32.89 & 28.08 & 18.35 & \textbf{30.42} & 32.36 & 28.26 & 22.40 & 41.85 \\ 
    OLMo-7B & 13.05 & 14.92 & 12.67 & 9.72 & 12.93 & 17.42 & 10.22 & 15.89 & 12.07 & 11.35 & 11.55 & 12.99 & 14.83 & 12.47 & 13.66  \\
    MAP-Neo-7B & 26.89 & 46.44 & 26.24 & 19.70 & 24.63 & 33.89 & 22.08 & 26.04 & 26.25 & 14.26 & 25.46 & 27.60 & 25.65 & 23.33 & 34.96 \\
    \midrule
    BaichuanSEED & 26.57 & 40.03 & 25.73 & 17.14 & 23.90 & 33.29 & 19.30 & 33.01 & 24.93 & 14.90 & 21.02 & 30.52 & 27.86 & 18.40 & \textbf{41.98} \\ \midrule \midrule
    Baichuan2-7B-Chat & 25.11 & 44.49 & 19.52 & 15.81 & 24.39 & 33.89 & 17.03 & 30.32 & 24.67 & 16.80 & 14.80 & 27.27 & 27.25 & 17.17 & 38.10 \\ 
    Baichuan2-13B-Chat & 28.03 & 45.47 & 25.86 & 16.78 & 25.37 & 38.51 & 15.17 & 30.68 & 31.23 & 18.71 & 19.62 & 32.47 & 30.06 & 19.25 & 43.23 \\ 
    Llama3-8B-Instruct & \textbf{41.88} & \textbf{66.53} & \textbf{37.14} & \textbf{32.24} & \textbf{41.46} & \textbf{53.20} & \textbf{31.37} & \textbf{47.68} & \textbf{40.16} & \textbf{25.34} & \textbf{35.16} & \textbf{33.05} & \textbf{40.48} & \textbf{33.49} & \textbf{58.02} \\
    Qwen1.5-7B-Chat & 16.29 & 28.45 & 21.80 & 11.13 & 16.34 & 13.86 & 10.42 & 20.78 & 11.29 & 11.08 & 20.50 & 13.64 & 23.05 & 10.24 & 15.54 \\ 
    OLMo-7B-Chat & 17.99 & 26.64 & 14.20 & 13.25 & 19.02 & 22.39 & 12.69 & 23.59 & 14.96 & 15.35 & 12.51 & 21.75 & 15.83 & 13.01 & 26.69 \\
    MAP-Neo-7B-Chat & 30.24 & 49.51 & 30.80 & 21.55 & 29.76 & 38.86 & 19.50 & 29.34 & 32.02 & 16.26 & 34.64 & 30.52 & 25.25 & 25.87 & 39.47 \\
    \midrule
    BaichuanSEED-SFT & 29.63 & 49.93 & 24.84 & 14.93 & 30.00 & 37.20 & 18.27 & 34.47 & 31.50 & 21.25 & 21.69 & 31.60 & 30.66 & 20.40 & 48.12 \\
    \bottomrule
    \end{tabular}}
    }
    \label{tab:mmlupro}
\end{table*}

\begin{table*}[htbp]
    \centering
    \small
    \caption{The detailed performance on LiveBench.}
    \setlength\tabcolsep{1mm}{
    \begin{tabular}{lccccccc}
    \toprule
    \textbf{Models} & \textbf{Avg.} & \textbf{Code} & \textbf{DA} & \textbf{IF} & \textbf{Language} & \textbf{Math} & \textbf{Reasoning}\\ \midrule
    Baichuan2-7B-Chat & 12.89 & 2.00 & 13.78 & 40.13 & 5.15 & 6.96 & 9.33 \\
    Baichuan2-13B-Chat & 13.04 & 2.00 & 11.20 & 44.72 & 4.70 & 6.94 & 8.67 \\
    Llama3-8B-Instruct & \underline{25.91} & \underline{16.95} & \underline{25.33} & \textbf{56.84} & \textbf{19.40} & 15.59 & \textbf{21.33} \\
    Qwen1.5-7B-Chat & 16.78 & 6.63 & 16.23 & 44.12 & 6.18 & 12.86 & 14.67 \\
    Qwen2-7B-Chat & \textbf{26.23} & \textbf{29.21} & \textbf{28.75} & 44.74 & \underline{10.21} & \textbf{25.83} & \underline{18.67} \\ 
    OLMo-7B-SFT & 8.80 & 0.00 & 7.33 & 32.76 & 3.53 & 5.81 & 3.33 \\ 
    MAP-Neo-7B-SFT & 14.35 & 7.95 & 4.27 & 36.84 & 7.48 & 11.59 & 18.00 \\ 
    \midrule
    BaichuanSEED-SFT & 18.32 & 5.00 & 15.75 & \underline{52.91} & 7.12 & \underline{16.46} & 12.67 \\
    \bottomrule
    \end{tabular}}
    \label{tab:livebench}
\end{table*}

The detailed performance of MMLU, CMMLU, and AGIEval can be found in Table~\ref{tab:mmlu}.
The detailed performance of each subjects of MMLU-Pro and LiveBench can be found in Table~\ref{tab:mmlupro} and Table~\ref{tab:livebench}, respectively.
All bold numbers indicate the best performance, while the underlined stand for the second.

\section{Case Study on Global Multi-granularity Deduplication}
\label{sec:case-study-deduplication-appendix}

To display the performance of our proposed global multi-granularity deduplication intuitively in Section~\ref{sec:reweighting}, we provide several cases before and after the deduplication, as shown in Figure~\ref{fig:dedup-cases}.

In terms of sentence deduplication, we filter from two dimensions: frequency and quality. 
On the one hand, reducing the frequency of repeated sentences, which may mitigate the repetition generation.
On the other hand, it can avoid low-quality sentences being learnt by the model. 
These two terms are indispensable and complementary to each other.
We conclude the characteristics of the removed sentences as follows, while filtering out these patterns may improve the accuracy and recall of harmful and toxic contents: 

\begin{itemize}
    \item \textbf{Advertisements.} SEO strategies are adopted by advertisements to increase visibility, such as creating spam sites and utilizing various trending long-tail keywords to mislead search engines. 
    The normal popular contents are interleaved with the advertisements, resulting in lots of template sentences. 
    Without deduplication, relying solely on the quality score is not sufficient to filter them out.
    \item \textbf{Texts for dissemination and interactivity.} It is widely utilized in web pages, which is one of the most common patterns. 
    While it is usually unrelated to the main idea of the documents.
    \item \textbf{Copyrights and PII.} These similar patterns mostly appear in news, books. 
    It may address privacy concerns, and aggravate the LLM repetition generation issue.
    \item \textbf{Residual HTML formats and watermarks.} Inappropriate custom parsing website templates may address this issue. 
\end{itemize}

\end{document}